# Optimize_Prime@DravidianLangTech-ACL2022: Abusive Comment Detection in Tamil

**Shantanu Patankar**\*, **Omkar Gokhale**\*, **Onkar Litake**†, **Aditya Mandke**†, **Dipali Kadam**
Pune Institute of Computer Technology, Pune, India
shantanupatankar2001@gmail.com, omkargokhale2001@gmail.com,
onkarlitake@ieee.org, adeetya.m@gmail.com, ddkadam@pict.edu

## Abstract

This paper tries to address the problem of abusive comment detection in low-resource indic languages. Abusive comments are statements that are offensive to a person or a group of people. These comments are targeted toward individuals belonging to specific ethnicities, genders, caste, race, sexuality, etc. Abusive Comment Detection is a significant problem, especially with the recent rise in social media users. This paper presents the approach used by our team — Optimize_Prime, in the ACL 2022 shared task "Abusive Comment Detection in Tamil." This task detects and classifies YouTube comments in Tamil and Tamil-English Codemixed format into multiple categories. We have used three methods to optimize our results: Ensemble models, Recurrent Neural Networks, and Transformers. In the Tamil data, MuRIL and XLM-RoBERTA were our best performing models with a macro-averaged f1 score of 0.43. Furthermore, for the Codemixed data, MuRIL and M-BERT provided sublime results, with a macro-averaged f1 score of 0.45.

## 1 Introduction

The rise in social media platforms like Facebook and Twitter has led to the exchange of massive amounts of information on the internet. With the increase in the number of users and platforms, problems like hate speech and cyberbullying have also increased (Chakravarthi, 2020). Abusive comments are comments that are offensive towards a particular individual or a group of individuals. Online abuse has led to problems like lowered self-esteem, depression, harassment, and even suicide in some severe cases. Hence detecting and dealing with such comments is of utmost importance. Classifying detected comments helps determine the severity of the comment and will also help the authorities take appropriate action against the individual. Our task is to detect and classify abusive comments written in Tamil. Abusive comment detection is a text classification problem. Text classification is a technique that extracts features from text and assigns a set of predefined categories(classes) to it. Traditionally, text classification was done using linear classifiers on the sentence embeddings of text. This was followed by Recurrent Neural Networks like LSTMs, which gave promising results. After the paper (Vaswani et al., 2017), transformers were introduced in the field of natural language processing. They have an attention layer that provides context to words in the text. The introduction of the transformer architecture has led to the development of many other variations of the transformer like BERT(Devlin et al., 2018), XLM-RoBERTa(Conneau et al., 2019), MuRIL(Khanuja et al., 2021), etc.

In this paper, we use different transformer-based models for abusive comment detection in Tamil. We have also used RNN models like LSTMs, a newer model ULMFit and a type of Ensemble model. We compared the results obtained from all three approaches to determine the optimum model for this task.

## 2 Related Work

Tamil is a low-resource language, so finding properly annotated data is challenging. In order to encourage research in Tamil, datasets have been created by Chakravarthi et al. (2020). The paper, Pitsilis et al. (2018) tries an RNN based approach for detecting offensive language in tweets. Arora (2020) developed a model for detecting hate speech in Tamil-English codemixed social media comments using a pre-trained version of ULM-FiT. After the introduction of transformers in Vaswani et al. (2017), the use of transformers for NLP tasks increased.

The release of BERT(Devlin et al., 2018) paved

---
\*first author, equal contribution
†second author, equal contribution

| Dataset | Hope-Speech | Homophobia | Misandry | Counter-Speech | Misogyny | Xenophobia | Transphobic | None-of-these |
|---|---|---|---|---|---|---|---|---|
| Tamil | 3.51% | 1.46% | 19.34% | 6.62% | 5.62% | 4.25% | 0.2% | 59% |
| Codemix | 3.61% | 2.91% | 14.4% | 5.71% | 3.42% | 4.97% | 2.74% | 62% |

Table 1: Distribution of classes in data

the way for many more variations of transformers. In the paper, Mishra and Mishra (2019) the results for the HASOC in Indo-European languages were showcased where they used MultiLingual BERT and monolingual BERT. Some work has been done by Ziehe et al. (2021) in English, Malayalam, and Tamil, aiming to detect Hope Speech which is also a text classification task. They fine-tuned XLM-RoBERTa (Conneau et al., 2019) for Hope Speech Detection.

The development of models like MuRiL (Khanuja et al., 2021), XLM-RoBERTa (Conneau et al., 2019), Indic BERT (Kakwani et al., 2020), and M-BERT (Devlin et al., 2018) has encouraged research in various low resource as well as high resource languages.

## 3 Dataset Description

The shared task on Abusive Comment Detection in Tamil-ACL 2022 aims to detect and reduce abusive comments on social media. The main objective of the shared task is to design systems to detect and classify instances of hate speech in Tamil and Tamil-English codemixed YouTube comments. The Abusive Comment Detection Dataset Priyadharshini et al. (2022) consists of Tamil and Tamil-English comments collected from the YouTube comments section. The dataset consists of a comment and its corresponding label belonging to the nine labels in the Dataset: Misandry, Counter-speech, Misogyny, Xenophobia, Hope-Speech, Homophobia, Transphobic, Not-Tamil, and None-of-the-above.

### 3.1 Tamil Data

The Train, Dev, and Test datasets have 2240, 560 and 700 data points, respectively. Each data point in the training data has the text in Tamil followed by its corresponding label.

### 3.2 Tamil-English Codemixed

The train, dev, and test datasets have 5948, 1488, and 1859 data points, respectively. Each data point has the actual comment in a codemixed format. Codemixed means text that alternates between two languages. In this case, the two languages are Tamil and English.

There is a significant class imbalance observed in the dataset. The 'Not-Tamil' label has no test or dev data instances, so the classification is done only for eight labels.

## 4 Methodology

To classify Youtube Comments, we used three different approaches: Ensemble models, Recurrent Neural Networks, and Transformers.[1]

### 4.1 Data Pre-processing

#### 4.1.1 Data cleaning

We removed punctuations, URL patterns, and stop words from the text. For better contextual understanding, we replaced emojis with their textual equivalents. For example, the laughing emoji was replaced by the Tamil equivalent of laughter.

Data cleaning boosted the performance of all RNN models and all Transformer models except for MuRIL. MuRIL and all Ensemble models worked better without data cleaning.

#### 4.1.2 Handling data imbalance

There is a significant class imbalance in the data. To reduce the class imbalance, we used the following techniques: over-sampling, over-under sampling, Synthetic Minority Over-sampling (SMOTE)(Chawla et al., 2002), and assigning class weights. In over-under sampling, we under-sample the classes having more instances than expected and over-sample those having lesser instances than expected while keeping the length of the dataset constant. Over-under sampling worked best for all transformer and ensemble models, but it reduced the performance of RNN models. Assigning weights boosted the performance of the M-BERT - Logistic Regression ensemble model.

---
[1]https://github.com/PICT-NLP/Optimize_Prime-DravidianLangTech2022-Abusive_Comment_Detection

## 4.2 Ensemble model

As shown in figure 1, we concatenate different machine learning models with multilingual BERT(M-BERT)(Devlin et al., 2018). Multilingual BERT is a BERT-based transformer trained in 104 languages. It simultaneously encodes knowledge of all these languages. M-BERT generates a sentence embedding vector of length 768. We then pass these embeddings to different machine learning models, as shown in Table 2. We used grid search with weighted-average f1 score as the scoring parameter for 5 -10 cross-validation folds to fine-tune the hyperparameters.

## 4.3 RNN Models

We have used two RNN models, Long Short-Term Memory(LSTM) networks and ULM-Fit.

### 4.3.1 Vanilla LSTM

Long Short-Term Memory (LSTM) (Hochreiter and Schmidhuber, 1997), can capture semantic information and long-term dependencies. We use LSTM to set a baseline score for RNN models. We create word embeddings by choosing the top 64,000 most frequently occurring words in the dataset. The embedding layer then creates 100-dimension vectors. The rest of the model includes a spatial drop out of 0.2, a single LSTM layer, and a final softmax activation function.

### 4.3.2 ULMFit

In transfer learning approaches, models are trained on large corpora, and their word embeddings are fine-tuned for specific tasks. In many state-of-the-art models, this approach is successful (Mikolov et al., 2013). Although Howard and Ruder (2018) argues that we should use a better approach instead of randomly initializing the remaining parameter. They have proposed ULMFiT: Universal Language Model Fine-tuning for Text Classification.

We use team gauravarora's (Arora, 2020) open-sourced models from the shared task at HASOC-Dravidian-CodeMix FIRE-2020. They build corpora for language modeling from a large set of Wikipedia articles.

These models are based on the Fastai (Howard and Gugger, 2020) implementation of ULMFiT. We tuned the models on Tamil, Codemix data sets individually and on the Tamil - codemix combined dataset.

For tokenization, we used the Senterpiece module. The language model is based on AWD-LSTM

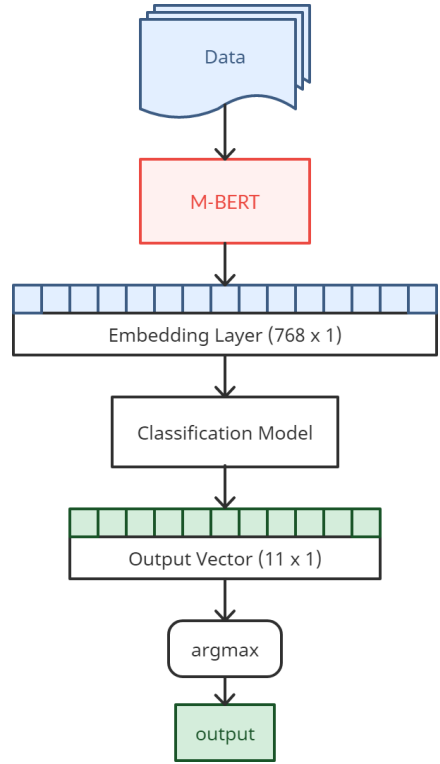

Figure 1: Ensemble model architecture

(Merity et al., 2018). The model consists of a regular LSTM cell with spatial dropout, followed by the classification model consisting of two linear layers followed by a softmax.

## 4.4 Transformer models

Our data sets consist of Tamil and Tamil-English codemixed data; we use four transformers MuRIL, XLM-RoBERTa, M-BERT, and Indic BERT. MuRIL (Khanuja et al., 2021) is a language model built explicitly for Indian languages and trained on large amounts of Indic text corpora. XLM-RoBERTa (Conneau et al., 2019) is a multilingual version of RoBERTa (Liu et al., 2019). Moreover, it is pre-trained on 2.5 TB of filtered CommonCrawl data containing 100 languages. M-BERT (Devlin et al., 2018) or multilingual BERT is pre-trained on 104 languages using masked language modeling (MLM) objective. Indic BERT (Kakwani et al., 2020) is a multilingual ALBERT (Lan et al., 2019) model developed by AI4Bharat and, it is trained on large-scale corpora of major 12 Indian languages, including Tamil. We use HuggingFace (Wolf et al., 2019) for training with SimpleTransformers. The training was stopped early if the f1 score did not improve for three consecutive epochs.

| Model Type | Classifier | Tamil | | Codemix | |
|---|---|---|---|---|---|
| | | macro f1 | weighted f1 | macro f1 | weighted f1 |
| Ensemble Models | Logistic Regression | 0.32 | 0.60 | 0.33 | 0.64 |
| | Decision Trees | 0.16 | 0.50 | 0.16 | 0.52 |
| | SVC | **0.33** | **0.61** | 0.32 | 0.64 |
| | Random Forest | 0.17 | 0.53 | 0.16 | 0.52 |
| | XG-Boost | 0.25 | 0.59 | 0.26 | 0.60 |
| | MLP | **0.33** | 0.60 | **0.35** | **0.65** |
| RNN Models | ULMFiT | 0.33 | **0.63** | **0.40** | <u>0.68</u> |
| | ULMFiT CD* | **0.36** | 0.61 | 0.38 | 0.62 |
| | Vanilla LSTM | 0.21 | 0.61 | 0.35 | 0.66 |
| Transformer Models | MuRIL | <u>**0.43**</u> | <u>0.68</u> | <u>**0.45**</u> | 0.60 |
| | XLM-R-base | <u>**0.43**</u> | 0.66 | 0.44 | **0.62** |
| | M-BERT | 0.40 | <u>**0.68**</u> | <u>**0.45**</u> | 0.61 |
| | Indic BERT | 0.40 | 0.65 | 0.35 | 0.52 |

Table 2: Results obtained by all the models on the Tamil as well as codemixed data.
*ULMFiT CD is trained by combining Tamil and CodeMix data.

## 5 Results

The results obtained by all the models can be viewed in Table 2.

### 5.1 Ensemble Models

In the case of the ensemble models, Support Vector Machine obtained the best result for the Tamil data. A macro-averaged f1 score of 0.33 and weighted-average f1 score of 0.6 was obtained.

In the case of the codemixed data, Multi-Layer Perceptron obtained the best score among the ensemble models. It achieved a macro-averaged f1 score of 0.35 and a weighted-average f1 score of 0.65.

Tree-based algorithms like decision trees, random forest, and XGBoost did not perform well.

### 5.2 RNNs

Among all the RNN models, ULMFiT fine-tuned on codemix data had the highest macro avg f1 of 0.40 and weighted avg f1 score of 0.68. ULMFit fine-tuned on Tamil data had the highest weighted avg f1 score of 0.63 while ULMFiT finetuned on combined dataset had the highest macro-averaged f1 score of 0.36.

### 5.3 Transformers

Out of the four transformers, we obtained the best results for MuRIL and XLM-RoBERTa in the case of Tamil. The macro-averaged f1 scores were 0.43 for both models, and the weighted-averaged f1 scores were 0.68 and 0.66 for MuRIL and XLM-RoBERTa, respectively.

MuRIL and M-BERT outperformed all the other models for the Tamil-English codemixed data. The macro-averaged f1 scores of 0.45 and weighted-average f1 scores of 0.61 were obtained by MuRIL and M-BERT.

## 6 Conclusion

This paper aims to detect and classify abusive comments. We tried three approaches for abusive comment detection in Tamil and Tamil-English Code-Mixed data: Ensemble models, Recurrent Neural Networks, and transformer-based models.

For the Tamil data, MuRIL and XLM-RoBERTa provided the best results with a macro-averaged f1 score of 0.43. Classes like Homophobia and Misandry were predicted with higher accuracy than others like Transphobic and Counter-Speech. Sentences that are not abusive are also classified well. For the codemixed data, MuRIL and M-BERT outperformed all other models with a macro-averaged f1 score of 0.45. Classes like Xenophobia, Misandry, and Transphobic were predicted with higher accuracy than others for the codemixed data. Sentences that are not abusive are also classified well.

In the future, various techniques can be tried to improve the performance of the models. In order to boost the performance, genetic algorithm-based ensembling methods could be used.